\PassOptionsToPackage{round,authoryear}{natbib}
\PassOptionsToPackage{table,dvipsnames}{xcolor}
\documentclass[]{style/company_light}

\usepackage[utf8]{inputenc}
\usepackage{hyperref}
\usepackage{url}
\usepackage{booktabs}
\usepackage{amsfonts}
\usepackage{amsmath}
\usepackage{nicefrac}
\usepackage{microtype}
\usepackage{graphicx}
\usepackage{subcaption}
\usepackage{multirow}
\usepackage{enumerate}
\usepackage{enumitem}
\usepackage[most]{tcolorbox}
\usepackage{adjustbox}
\usepackage{array}
\usepackage{amssymb}

\RequirePackage{colortbl}
\colorlet{tabletotal}{brandbg}
\newcommand{\deltasr}[1]{\textcolor{brandaccent}{#1}}

\graphicspath{{figures/}}

\newcommand{\methodname}{\textsc{BrowserBC}}
\newcommand{\keyclaim}[1]{\textcolor{brandaccent}{\emph{#1}}}

\title{Scalable Behaviour Cloning on Browser Using via Skill Distillation}

\abstract{%
{\bfseries
Internet users collectively perform an enormous range of skilled work through web browsers, from software development and document editing to search, forms, and enterprise workflows, making human browsing a highly scalable but under-exploited source of reusable browser skills. We argue that the bottleneck for browser agents is decision-making under incomplete information rather than low-level operation, and that the priors agents lack are already implicit in human interaction traces. We therefore study scalable behavior cloning for browser agents via skill distillation, converting user interaction trajectories into compact natural-language skills that agents can read, retrieve, reuse, and compose directly. We further organize the distilled skills into a skill graph so that growth proceeds through consolidation rather than unbounded accumulation. This suggests that the scalability of browser agents may come less from manually designed tasks and more from the collective skills already expressed by internet users. Our project is available at: \url{https://lab.einsia.ai/browserbc/}.
}
}

\date{\today}

\begin{document}

\maketitle

\section{Introduction}

Web browsers have become a universal interface for digital work: developers navigate code hosting platforms and documentation, operators monitor enterprise dashboards, researchers collect information across online tools, and everyday users manage email, calendars, and software-as-a-service applications. A browser agent that can reliably operate this interface would therefore serve as a general execution layer for digital labor. This vision has motivated steady progress, from early demonstrations of language agents acting in grounded web environments \citep{yao2023webshopscalablerealworldweb,deng2023mind2webgeneralistagentweb} to increasingly realistic benchmarks that require long-horizon interaction, multimodal observation, and element grounding on real websites \citep{zhou2024webarenarealisticwebenvironment,koh2024visualwebarenaevaluatingmultimodalagents,drouin2024workarenacapablewebagents,dechezelles2025browsergymecosystemwebagent}.

Despite this progress, contemporary web agents remain inefficient on real websites. We argue that the central bottleneck lies not in low-level browser operation but in decision-making under incomplete information: modern agents can usually parse rendered or structured observations and execute clicks, keystrokes, and submissions, so the hard question is not \emph{whether} an action can be executed but \emph{which} action should be taken. Because an agent observes only the current page while the global site structure, reliable entry points, path costs, and completion signals remain hidden, even capable agents are forced into trial-and-error exploration---a symptom of weak decision-making under uncertainty rather than insufficient motor control, aligning with prior evidence that web interaction requires planning and long-context reasoning beyond flat action imitation \citep{gur2024realworldwebagentplanninglong,zheng2024gpt4visiongeneralistwebagent,xie2024osworldbenchmarkingmultimodalagents}.

\begin{figure}[t]
\centering
\includegraphics[width=1\textwidth]{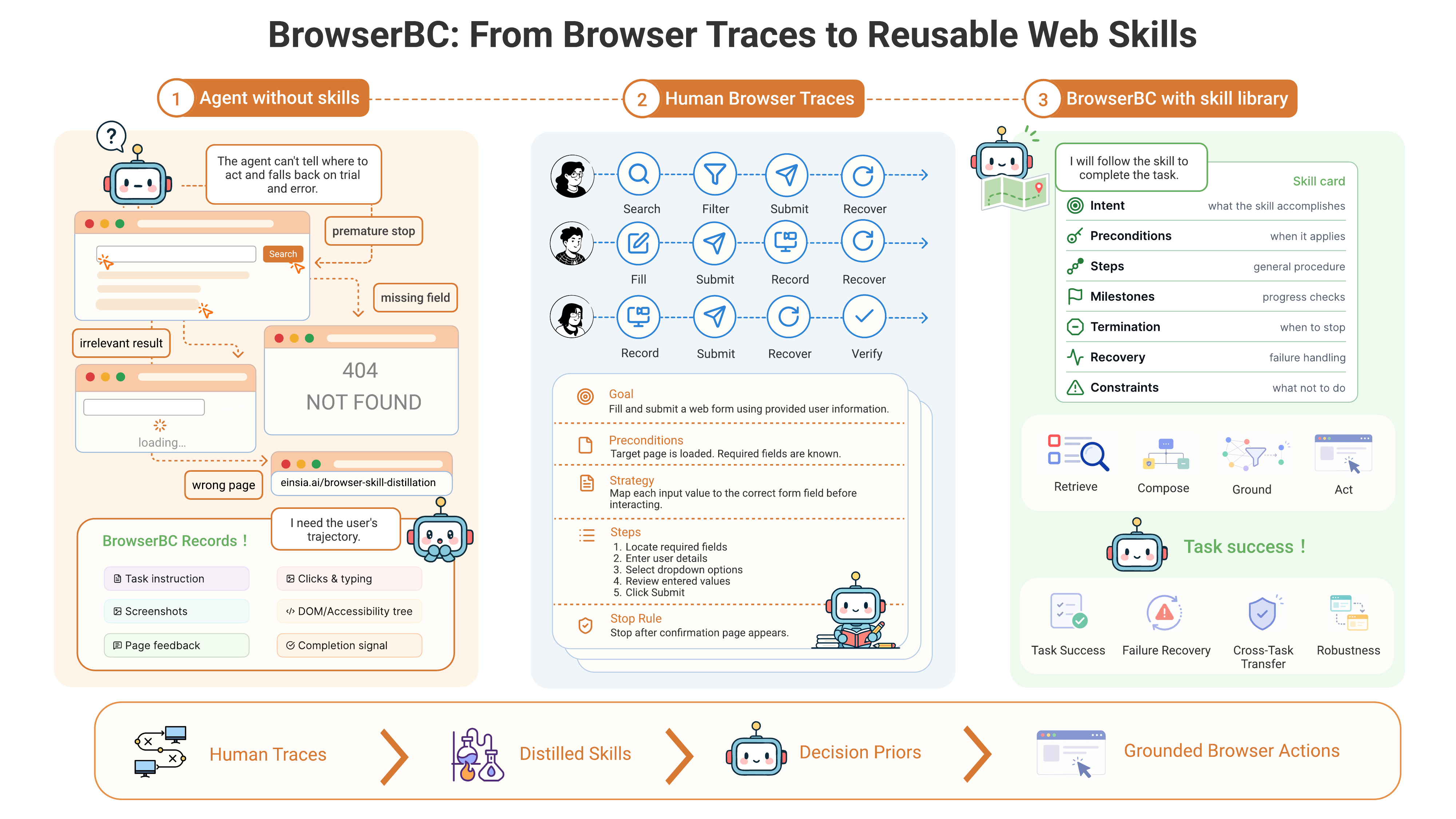}
\caption{\textbf{Motivation of BrowserBC.} \methodname{} turns human browsing traces into reusable natural-language skills, providing agents with decision priors for navigating unfamiliar websites.}
\label{fig:introduction}
\end{figure}

Crucially, the decision priors missing from current agents are abundant in human behavior. People complete browser tasks every day without solving each one from scratch, relying on implicit knowledge that is rarely stated explicitly on the page but is revealed through the paths they take. Browser traces are therefore more than action logs: they are large-scale records of human decision-making under partial observability, and recent datasets and systems have made such traces increasingly accessible, including demonstrations for grounding and next-step prediction, dialogue-conditioned navigation traces, and trajectories synthesized from web tutorials \citep{deng2023mind2webgeneralistagentweb,lù2024weblinxrealworldwebsitenavigation,xu2025agenttrekagenttrajectorysynthesis}.

However, raw traces are a poor unit of reusable supervision. They are long, noisy, redundant, and tightly coupled to incidental page states. Their low-level targets often depend on coordinates, DOM structure, transient identifiers, login state, and page versions. Directly cloning clicks or next actions therefore risks learning brittle, page-specific behavior rather than transferable competence. We argue that the reusable unit of browser behavior is neither an individual action nor an entire trajectory, but a transferable natural-language \emph{skill}: a structured description of a subtask strategy that specifies its intent, applicability, preconditions, execution procedure, progress milestones, termination evidence, and failure-recovery steps.

This skill-centric view builds on evidence that language agents benefit from persistent verbal experience, including textual feedback, reusable lessons from prior trials, and skill libraries that abstract beyond raw action sequences \citep{shinn2023reflexionlanguageagentsverbal,zhao2024expelllmagentsexperiential,wang2023voyageropenendedembodiedagent,wang2025inducingprogrammaticskillsagentic}. Representing experience as natural language offers a key advantage: the producer of a skill can be decoupled from its executor. A capable human or model may complete a task once; the resulting trajectory can then be distilled into a reusable skill that a smaller or cheaper model reads and applies to related tasks. In this way, experience recorded once can be shared across agents, offering a scalable path to stronger web browsing without requiring every agent to be trained or scaled independently.

Building on these observations, we propose \textbf{\methodname{}}, a scalable framework that converts browser interaction traces into a retrievable, composable, and executable library of natural-language skills. \methodname{} first cleans and semantically segments each trace, then distills the resulting evidence into structured skill cards that preserve transferable procedural knowledge while removing volatile or task-leaking details. Candidate skills are further organized into a skill graph, allowing the library to grow through consolidation rather than unbounded accumulation. At inference time, a lightweight retriever selects a small set of relevant skills as context, and the agent grounds them into element selection and executable actions on the live page.

\paragraph{Contributions.} This paper makes the following contributions:
\begin{itemize}
\item We reframe the inefficiency of capable web agents as a decision-making problem under incomplete information, and identify human browser traces as large-scale decision data whose reusable unit is a transferable natural-language skill rather than a click or a full trajectory.
\item We propose \methodname{}, a scalable framework that distills browser traces into structured skill cards, consolidates them into a skill graph, and decouples skill provenance from skill execution so that experience recorded once can be reused across models.
\item We evaluate \methodname{} on challenging web-agent benchmarks, measuring next-action prediction, end-to-end task success, failure recovery, skill reuse, cross-site generalization, and robustness to page perturbations.
\end{itemize}
\section{Related Work}

\paragraph{Web and computer-use agents.}
Research on browser agents has advanced along three coupled fronts: environments, benchmarks, and end-to-end policies. Early work established grounded language-agent interaction in controlled settings, from WebShop in a simulated storefront \citep{yao2023webshopscalablerealworldweb} to Mind2Web over real websites \citep{deng2023mind2webgeneralistagentweb}. WebArena and VisualWebArena subsequently made evaluation realistic by requiring long-horizon interaction, multimodal perception, element grounding, and programmatic success checks \citep{zhou2024webarenarealisticwebenvironment,koh2024visualwebarenaevaluatingmultimodalagents}, and WorkArena and BrowserGym extended this line toward enterprise knowledge work and a shared research infrastructure \citep{drouin2024workarenacapablewebagents,boisvert2025workarenacompositionalplanningreasoningbased,dechezelles2025browsergymecosystemwebagent}. A parallel effort strengthens the policy itself: WebVoyager and OpenWebVoyager study multimodal agents under realistic browsing \citep{he2024webvoyagerbuildingendtoendweb,he2024openwebvoyagerbuildingmultimodalweb}, SeeAct isolates grounding as the central bottleneck for vision-language agents \citep{zheng2024gpt4visiongeneralistwebagent}, and WebAgent couples planning, long-context reading, and program synthesis \citep{gur2024realworldwebagentplanninglong}. Beyond the browser, OSWorld generalizes the problem to open-ended computer use and argues for robust observation--action abstractions over narrow memorization of UI state \citep{xie2024osworldbenchmarkingmultimodalagents}, while large-scale efforts such as ScaleCUA and OpenCUA characterize how competence scales with data and supervision \citep{liu2026scalecua,wang2025opencua}. More recent systems push capability through reinforcement learning and information-seeking objectives \citep{li2026websailorv2,wu2025webdancer,vattikonda2025trainwebagent}, process reward modeling \citep{chae2025webshepherd}, structured or hierarchical exploration \citep{yang2025sage,gandhi2026gobrowse,shen2025thinkingdoing}, and tool or context abstractions \citep{prabhu2026walt,ye2026agentfold}. The recurring theme is that browser competence improves with scale, abstraction, and richer interaction. Our work is orthogonal to all three fronts: rather than proposing new environments, rewards, or training objectives, we treat the trajectories already produced by ordinary internet users as a scalable supervision signal and ask how to convert them into reusable procedural knowledge.

\paragraph{Learning from web demonstrations.}
A substantial body of work uses human or agent trajectories as supervision. Mind2Web shows that demonstrations over real websites can supervise action grounding and next-step prediction \citep{deng2023mind2webgeneralistagentweb}; WebLINX collects multi-turn, dialogue-conditioned navigation traces from real-world use \citep{lù2024weblinxrealworldwebsitenavigation}; and AgentTrek synthesizes trajectories from web tutorials as an alternative route to scalable data \citep{xu2025agenttrekagenttrajectorysynthesis}. Such traces implicitly encode how people decompose work, locate pages, select elements, recover from mistakes, and recognize task completion. The central difficulty is one of representation: raw traces are long, noisy, and tightly coupled to incidental page state, so directly cloning coordinates, selectors, or next-step actions tends to capture brittle, page-specific correlations rather than transferable competence. We share the goal of exploiting demonstrations at scale, but we take the reusable unit of supervision to be a transferable natural-language \emph{skill} rather than a single action or an entire trajectory, and we deliberately abstract away coordinates and selectors during distillation so that the resulting guidance survives changes in page version and authentication state.

\paragraph{Skills and reusable experience.}
Our formulation builds on evidence that language agents benefit from persistent verbal experience. Reflexion converts outcome feedback into textual self-notes that improve later attempts \citep{shinn2023reflexionlanguageagentsverbal}, ExpeL extracts reusable lessons across trials \citep{zhao2024expelllmagentsexperiential}, and Voyager maintains an expanding library of executable skills for open-ended embodied exploration \citep{wang2023voyageropenendedembodiedagent}. Related work induces programmatic skills that represent behavior above the level of raw action sequences \citep{wang2025inducingprogrammaticskillsagentic}, while modular frameworks such as Agent Lumos decouple planning, grounding, and execution into reusable components \citep{yin2024agentlumosunifiedmodular}. A complementary observation is that recorded experience can be reused beyond the agent that produced it: memory-sharing schemes pool useful experiences across language agents \citep{gao2024memorysharing}, and robotics efforts such as Open X-Embodiment show that heterogeneous demonstrations collected across platforms can support transfer \citep{openxembodiment2024}. We differ from these lines in both source and representation: we distill \emph{human} browser trajectories spanning many users, websites, and tasks into an auditable skill library, and---because each skill is a self-contained natural-language description---we decouple the model that \emph{produces} a skill from the model that \emph{executes} it, so that experience recorded once can be reused across different and cheaper models, with many trajectories contributing to one skill and one trajectory yielding several skills.

\section{Method}
\label{sec:method}

\methodname{} is a \emph{skill-centric} behavior cloning framework that distills human browser trajectories into reusable natural-language \emph{skills}, rather than imitating demonstrations as flat sequences of low-level actions. Each skill encodes a transferable procedure for a class of browser tasks: when it applies, which information must be preserved across steps, how progress is made, how completion is verified, and how common failures are diagnosed and recovered from. We first motivate this design (\S\ref{sec:motivation}) and formalize the skill-centric view (\S\ref{sec:formulation}), and then present the four stages of the framework: behavioral evidence abstraction (\S\ref{sec:evidence}), procedural skill distillation (\S\ref{sec:distill}), skill-library construction (\S\ref{sec:library}), and skill-conditioned execution (\S\ref{sec:execution}).

\begin{figure}[t]
\centering
\includegraphics[width=1\textwidth]{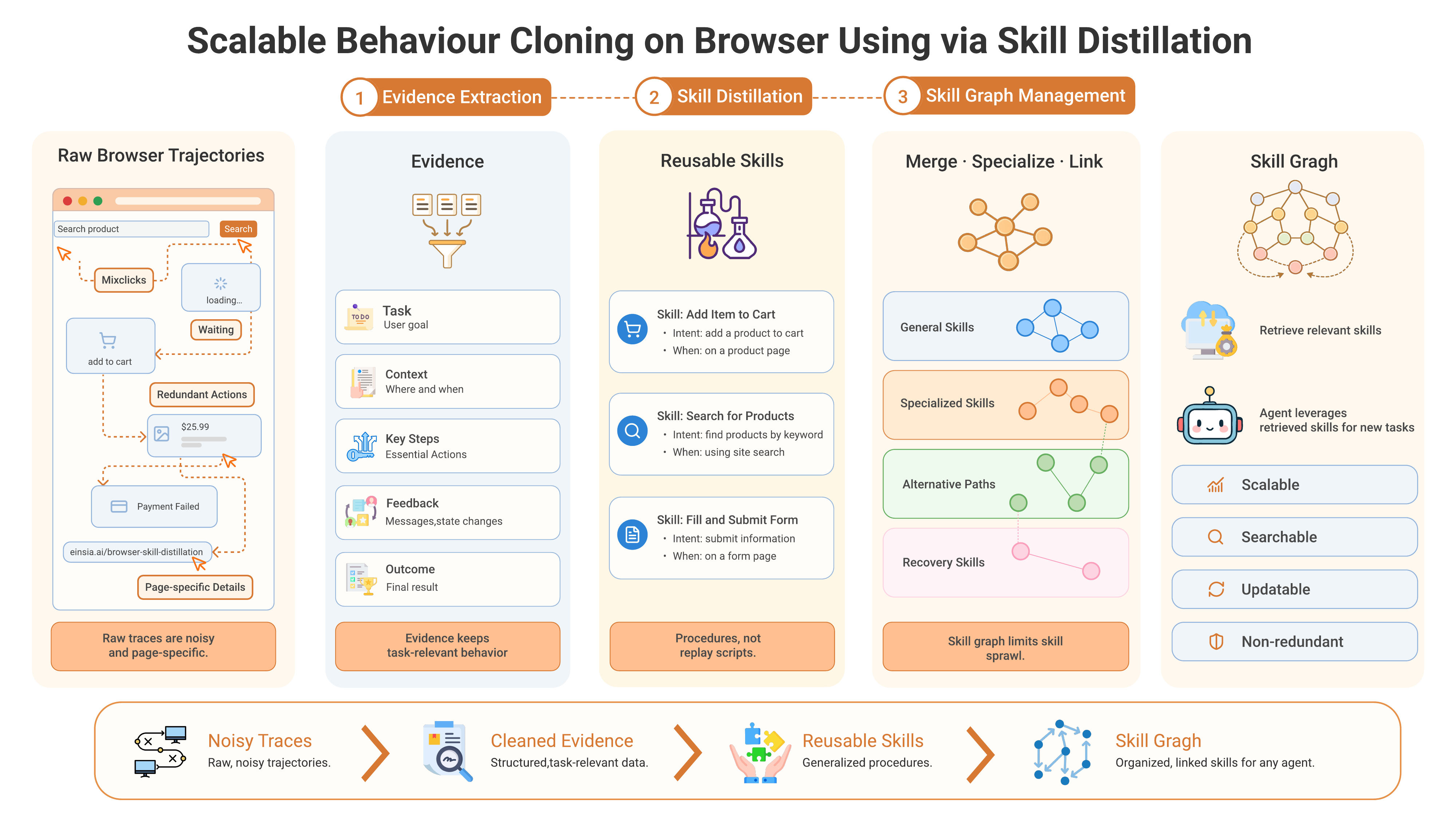}
\caption{\textbf{Overview of \methodname{}.} \methodname{} cleans and segments raw browser traces, distills them into structured natural-language skills, organizes the skills into a graph, and retrieves relevant skills to guide browser-agent execution on new tasks.}
\label{fig:overview}
\end{figure}

\subsection{Motivation}
\label{sec:motivation}

Modern web agents can already perceive rendered or structured pages and execute clicks, keystrokes, and submissions, yet they remain inefficient on real websites because they must act under incomplete information: an agent observes only the current page, while the global site structure, reliable entry points, relative path costs, and completion signals stay hidden. Lacking these decision priors, even strong agents fall back on trial-and-error exploration---entering plausible but irrelevant pages, oscillating among search results, halting before reaching the answer, or omitting critical fields in long forms. Such exploration is especially costly on long-horizon tasks and on websites with unfamiliar or deeply nested structure, where a single task may require many redundant interactions before it succeeds. The bottleneck is therefore the absence of priors about \emph{where to go and when to stop}, not a deficit in low-level browser operation.

Such priors, however, are abundant in human behavior: people complete browser tasks every day without re-deriving each one from scratch, relying on implicit knowledge of where to begin, how to refine search results, which form fields matter, and what feedback indicates progress or completion. This motivates cloning behavior from human traces---but cloning the \emph{right} unit. Imitating low-level actions is brittle on the web, where coordinates, DOM structures, layouts, and login state change constantly, so a replayed action breaks as soon as the page shifts. We instead clone the procedural knowledge \emph{behind} the actions, distilling each trajectory into a natural-language skill that specifies when it applies, how to navigate and verify progress, when to terminate, and how to recover from failure. Because a skill is a self-contained natural-language object, it can be retrieved, composed, and reused across websites, and the model that distills a skill need not be the one that executes it---so competence demonstrated once can be shared across smaller and cheaper agents.

\subsection{Problem Formulation}
\label{sec:formulation}

We denote a browser trajectory as $\tau = (x,\, e_1, \ldots, e_T,\, y)$, where $x$ is a natural-language task instruction, $e_t = (o_t,\, a_t,\, f_t)$ is the $t$-th event comprising the observation $o_t$, the user action $a_t$, and the execution feedback $f_t$ (e.g., page transitions, validation messages, or completion signals), and $y$ is the task outcome when available. A standard behavior cloning objective fits a low-level policy $\pi(a_t \mid x,\, o_{\leq t})$ that maps the task and interaction history directly to the next action; as argued above, such raw-action targets capture page-specific correlations rather than the reusable strategy that explains \emph{why} a demonstration succeeds.

We instead cast behavior cloning over a \emph{skill library} $\mathcal{L} = \{s_i\}_{i=1}^{N}$, where each $s_i$ is a natural-language procedure distilled from one or more trajectories. Given a new instruction $x^\star$ and browser history $h_t^\star = o^\star_{\leq t}$, the agent retrieves a compact, task-relevant subset of skills and predicts the next action through a \emph{skill-conditioned} policy:
\begin{equation}
    \mathcal{R}_t = \mathcal{R}\!\left(x^\star,\, h_t^\star,\, \mathcal{L}\right),
    \quad |\mathcal{R}_t| \ll |\mathcal{L}|,
    \qquad
    a_t \sim \pi_\theta\!\left(\cdot \mid x^\star,\, h_t^\star,\, \mathcal{R}_t\right).
    \label{eq:bc-skill}
\end{equation}
The retrieved skills $\mathcal{R}_t$ do not replace perception or grounding; they supply \emph{procedural constraints} on what the agent should attend to, which values it must preserve, what intermediate progress looks like, when a task should terminate, and how to recover when a step fails. Crucially, the library $\mathcal{L}$ is decoupled from the policy $\pi_\theta$ that consumes it: skills are distilled once and serve as model-agnostic guidance, so the executing agent can differ from---and be substantially cheaper than---the one used during distillation.

\subsection{Behavioral Evidence Abstraction}
\label{sec:evidence}

\keyclaim{Challenge 1: Raw trajectories must be cleaned and abstracted before any skill can be distilled from them.} They are noisy and heterogeneous, interleaving misclicks, idle waits, repeated attempts, and transient page states with the few interactions that actually drive task progress; they are also saturated with volatile or privacy-sensitive content---exact coordinates, DOM selectors, transient identifiers, login state, and personal text---that is irrelevant or even harmful to retain. Distilling skills from such logs without first isolating the task-relevant signal would propagate noise and non-transferable artifacts into every downstream skill.

The first stage therefore converts raw trajectories into structured \emph{evidence}. We normalize each trajectory into a common event representation, filter out interactions that do not contribute to task progress, and segment the remainder into semantically coherent sub-procedures---rather than fixed-length windows---so that each segment corresponds to a meaningful unit of behavior. For each segment we extract an \emph{evidence unit}
\begin{equation}
    u = \big(x,\; c^{\text{pre}},\; b,\; c^{\text{post}},\; f,\; y\big),
    \label{eq:evidence}
\end{equation}
comprising the instruction $x$, the contexts $c^{\text{pre}}$ and $c^{\text{post}}$ summarizing the state immediately before and after the segment, a summary $b$ of the demonstrated behavior, the feedback $f$ observed during the segment, and the outcome $y$. The context summaries retain task-relevant information---visible affordances, required input fields, user-provided values, applicable constraints, and success or failure signals---while discarding non-transferable artifacts, lifting demonstrations from page-specific action traces into procedural evidence that generalizes across browser states.

\subsection{Procedural Skill Distillation}
\label{sec:distill}

\keyclaim{Challenge 2: Even on clean evidence, cloning behavior as a replayable trace is fundamentally brittle on the web.} Layouts, DOM structures, page versions, and authentication states change constantly, so a procedure pinned to specific coordinates or selectors breaks as soon as the page shifts; meanwhile the genuinely transferable content of a demonstration---\emph{what} to accomplish and \emph{how} to tell that it is progressing and complete---is never made explicit by a raw action sequence. The challenge is to extract this reusable procedural knowledge while deliberately stripping away the volatile and leak-prone details that do not transfer.

Given a set of evidence units $\mathcal{U}_s$ and optional metadata $m_s$ (e.g., task category or website family), \methodname{} distills an agent-readable skill $\hat{s} = D(\mathcal{U}_s,\, m_s)$. A skill is a structured natural-language \emph{card} with a fixed set of fields---intent, applicable scope, preconditions, execution steps, progress milestones, terminal evidence, failure modes, recovery strategies, negative constraints, and provenance---which make it directly consumable by a language-model agent while keeping it human-auditable and easy to update. The distillation model is instructed to retain reusable procedural structure while removing session-specific artifacts, and a leakage audit ensures that no task-specific answer or evaluation-checker content is silently written into a card unless it is genuinely essential to the task semantics. A distilled skill thus specifies \emph{what} should be accomplished and \emph{how} to reason about progress, not how to replay a particular demonstration: a form-filling skill, for instance, instructs the agent to identify fields by their semantic labels and preserve task-provided values, rather than to reuse a coordinate that happened to work in the source trajectory. Skills may be induced from a single successful trajectory---which already captures one viable solution---or consolidated from multiple attempts, where successful runs strengthen the execution guidance and failed runs expose missing preconditions and motivate explicit recovery strategies.

\subsection{Skill-Library Construction}
\label{sec:library}

\keyclaim{Challenge 3: Distilling one skill per trajectory does not by itself yield a scalable system.} If every demonstration produces an independent entry, the library quickly degenerates into a redundant, ever-growing pile of overlapping and sometimes conflicting skills, in which retrieval becomes diffuse and any update risks touching unrelated entries. Scalability requires that the library be \emph{compressed and structured} as it grows, not merely accumulated.

To this end, whenever a candidate skill $\hat{s}$ is produced, \methodname{} decides whether to \emph{add} it as a new entry, \emph{merge} it into an existing skill, or register it as a \emph{specialization} of a more general one. Two skills are merged when they share a compatible intent, preconditions, steps, effects, and terminal evidence, and are kept separate when they apply under different conditions, require different information, or encode conflicting constraints. We organize the resulting library as a lightweight \emph{skill graph} $\mathcal{G} = (\mathcal{L}, \mathcal{E})$, whose nodes are distilled skills and whose edges encode simple relations---temporal dependency, specialization, alternative procedures for a shared subgoal, or incompatibility under the same state---so that a generic procedure (e.g., form filling) can link to its specializations (such as payment or profile updates) and to the recovery skills that handle its characteristic failures. This graph serves as the organizing scaffold for the skills: the unit that is learned and reused remains the distilled natural-language skill card, while the edges keep storage, retrieval, and revision local. As a result, \methodname{} consolidates repeated demonstrations into reusable nodes rather than isolated examples, restricts retrieval and updates to the relevant region of the skill space, and supports incremental refinement, since a new trajectory updates only the affected skills and their immediate neighbors.

\subsection{Skill-Conditioned Execution}
\label{sec:execution}

At inference time, the retrieval operator $\mathcal{R}$ in Eq.~\eqref{eq:bc-skill} selects the skills most relevant to the task by semantic similarity and, when available, compatibility with the current browser context; we deliberately avoid a specialized retrieval architecture or an explicit symbolic planner, so retrieval simply exposes the agent to the prior knowledge most likely to be useful at the current step. The selected skills $\mathcal{R}_t$ are inserted into the agent's context as natural-language guidance, and the agent grounds them into executable actions against the live page.

Skills are thus neither executable scripts nor demonstrations to be replayed verbatim: they bias the agent toward distilled behavior patterns, while each concrete action is still chosen with respect to the current state. When applying a form-filling skill, for example, the agent identifies the required fields by their labels, transfers the task-provided values into them, monitors for validation messages, and stops once a success confirmation appears---reading the concrete fields and layout from the current page rather than copying them from any source trajectory. Because this guidance is purely textual and model-agnostic, the executing agent may be a lighter-weight model than the one used for distillation, preserving the data efficiency of behavior cloning while avoiding the brittleness of action replay and allowing a single skill to guide behavior across different websites, interaction mechanisms, and underlying models.
\section{Experiments}
\label{sec:experiments}

\subsection{Experimental Setup}
\label{sec:exp-setup}

We evaluate \methodname{} on three benchmarks that probe complementary axes of browser- and computer-use competence, progressing from controlled reproducibility, to real-world volatility, and finally to cross-substrate generalization (Table~\ref{tab:benchmarks}). WebArena-Hard~\citep{zhou2024webarenarealisticwebenvironment} comprises $258$ human-verified tasks across six self-hosted sites with original programmatic checkers; its stable, reproducible environment makes it our primary setting for controlled comparison and mechanistic analysis. ClawBench complements this with tasks on \emph{live production websites}, combining layout volatility, write-heavy state changes, and human-grounded agentic verification. Because the same page may change between runs, ClawBench constitutes our strongest test of whether \methodname{} skills encode transferable procedural knowledge rather than brittle, page-specific action traces. Finally, OSWorld-style desktop tasks move interaction off the browser and onto applications, files, dialogs, and persistent system state, serving as a diagnostic study of whether the skill abstraction transfers across interaction substrates.

\begin{table}[h]
\centering
\caption{Summary of the three evaluation benchmarks. The three settings span controllability, real-world fidelity, and cross-substrate generalization, jointly testing whether \methodname{} clones reusable procedural knowledge rather than surface actions.}
\label{tab:benchmarks}
\begin{adjustbox}{max width=\linewidth}
\begin{tabular}{llll}
\toprule
Benchmark & Environment  & Verification & Primary stress axis \\
\midrule
WebArena-Hard & Self-hosted web & Programmatic checkers & Controllability / reproducibility \\
ClawBench & Live production websites & Agentic pass/fail evaluator & Real-world volatility / write-heavy state \\
OSWorld  & Ubuntu desktop apps & State-based verifiers & Cross-substrate generalization \\
\bottomrule
\end{tabular}
\end{adjustbox}
\end{table}

Unless otherwise stated, both the skill distiller and the executing agent are instantiated with Claude-Sonnet-4.6 under a fixed action interface and step budget. The only exception is the distiller--executor decoupling study (\S\ref{sec:exp-analysis}), which deliberately varies the distilling and executing models to test model-agnosticism. All other comparisons hold the agent, action interface, and step budget fixed, varying only whether retrieved \methodname{} skills are inserted into the agent's context. To control leakage, skills are restricted to procedural content, and any checker-specific or answer-bearing values are audited and redacted before evaluation; identical checker normalizations are applied to the skill-off and skill-on runs.

\subsection{Main Results}
\label{sec:exp-main}

Table~\ref{tab:main-results} reports task-group results on WebArena-Hard and per-category results on ClawBench for the identical browser agent with skills off (Base) and with the \methodname{} skill library.

\begin{table*}[ht]
\centering
\caption{Performance comparison across WebArena-Hard and ClawBench.
All Base and \methodname{} results are reported as success rates (SR, \%), and $\Delta$ SR denotes the absolute improvement in success rate.}
\label{tab:main-results}

\renewcommand{\arraystretch}{1.08}
\setlength{\tabcolsep}{10pt}
\footnotesize

\begin{adjustbox}{max width=\textwidth}
\begin{tabular}{lcccc}
\toprule
Task group / category & \# Cases & Base (Skill-Off) & \methodname{} & $\Delta$ SR \\
\midrule

\multicolumn{5}{l}{\sffamily\bfseries\color{brandaccent}WebArena-Hard} \\
\cmidrule(lr){1-5}
GitLab           & 57  & 64.9 &  86.0 & \deltasr{+21.1} \\
Shopping         & 56  & 60.7 &  89.3 & \deltasr{+28.6} \\
Shopping (admin) & 55  & 56.4 &  70.9 & \deltasr{+14.5} \\
Reddit           & 42  & 78.6 &  85.7 & \deltasr{+7.1}  \\
Multi-site       & 48  & 43.8 &  75.0 & \deltasr{+31.2} \\
\rowcolor{tabletotal}
\textbf{All}     & \textbf{258} & \textbf{60.5} & \textbf{81.4} & \textbf{\deltasr{+20.9}} \\

\addlinespace[0.45em]
\midrule

\multicolumn{5}{l}{\sffamily\bfseries\color{brandaccent}ClawBench} \\
\cmidrule(lr){1-5}
Daily    & 57  & 24.6 &  64.9 & \deltasr{+40.3} \\
Finance  & 6   & 50.0 & 100.0 & \deltasr{+50.0} \\
Work     & 17  & 47.1 &  76.5 & \deltasr{+29.4} \\
Dev      & 18  & 33.3 &  66.7 & \deltasr{+33.4} \\
Academic & 14  & 50.0 &  78.6 & \deltasr{+28.6} \\
Travel   & 13  & 38.5 &  76.9 & \deltasr{+38.4} \\
Social   & 16  & 25.0 &  56.2 & \deltasr{+31.2} \\
Pets     & 11  & 27.3 &  54.5 & \deltasr{+27.2} \\
\rowcolor{tabletotal}
\textbf{All} & \textbf{152} & \textbf{32.9} & \textbf{68.4} & \textbf{\deltasr{+35.5}} \\

\bottomrule
\end{tabular}
\end{adjustbox}
\end{table*}

\paragraph{WebArena-Hard.}
\methodname{} raises overall task success from $60.5\%$ ($156/258$) to $81.4\%$ ($210/258$), a $+20.9$-point absolute gain aggregated across task groups. The improvement is largest on Multi-site and Shopping tasks, where procedural navigation, multi-step search, and result interpretation are most demanding. In contrast, the smaller gains on Shopping-admin and Reddit indicate that skills help most when the bottleneck is procedural uncertainty rather than low-level execution fidelity.

\paragraph{ClawBench.}
On ClawBench, the skill-free baseline solves $50/152$ tasks ($32.9\%$), closely matching the difficulty profile reported for frontier agents, whereas \methodname{} solves $104/152$ tasks ($68.4\%$)---a $+35.5$-point absolute improvement that roughly doubles the number of solved tasks and is broadly distributed across all eight categories. This gain is notable precisely because ClawBench removes the conditions under which a skill library could succeed through memorization: since tasks run on live websites whose layout, availability, and flows change between runs, any skill encoding exact coordinates, DOM selectors, or cached page state would degrade rather than help. The large improvement therefore indicates that \methodname{} skills transfer at the level of \emph{what to accomplish and how to verify completion}. This interpretation is further consistent with ClawBench's write-heavy emphasis: skills that supply explicit preconditions, the state-changing operation, and the terminal evidence confirming submission directly address the completion-recognition failures that the agentic evaluator penalizes. Finally, the magnitude of this gain on a live, unsaturated benchmark---larger in relative terms than the WebArena gain---suggests that procedural priors are most valuable where the base agent's bottleneck is procedural uncertainty on unfamiliar real-world sites rather than low-level grounding.

\subsection{Analysis and Ablations}
\label{sec:exp-analysis}

\begin{figure}[t]
\centering

\newlength{\figH}
\setlength{\figH}{0.31\linewidth}

\begin{subfigure}[t]{0.58\linewidth}
    \centering
    \includegraphics[
        width=1.7778\figH,
        height=\figH
    ]{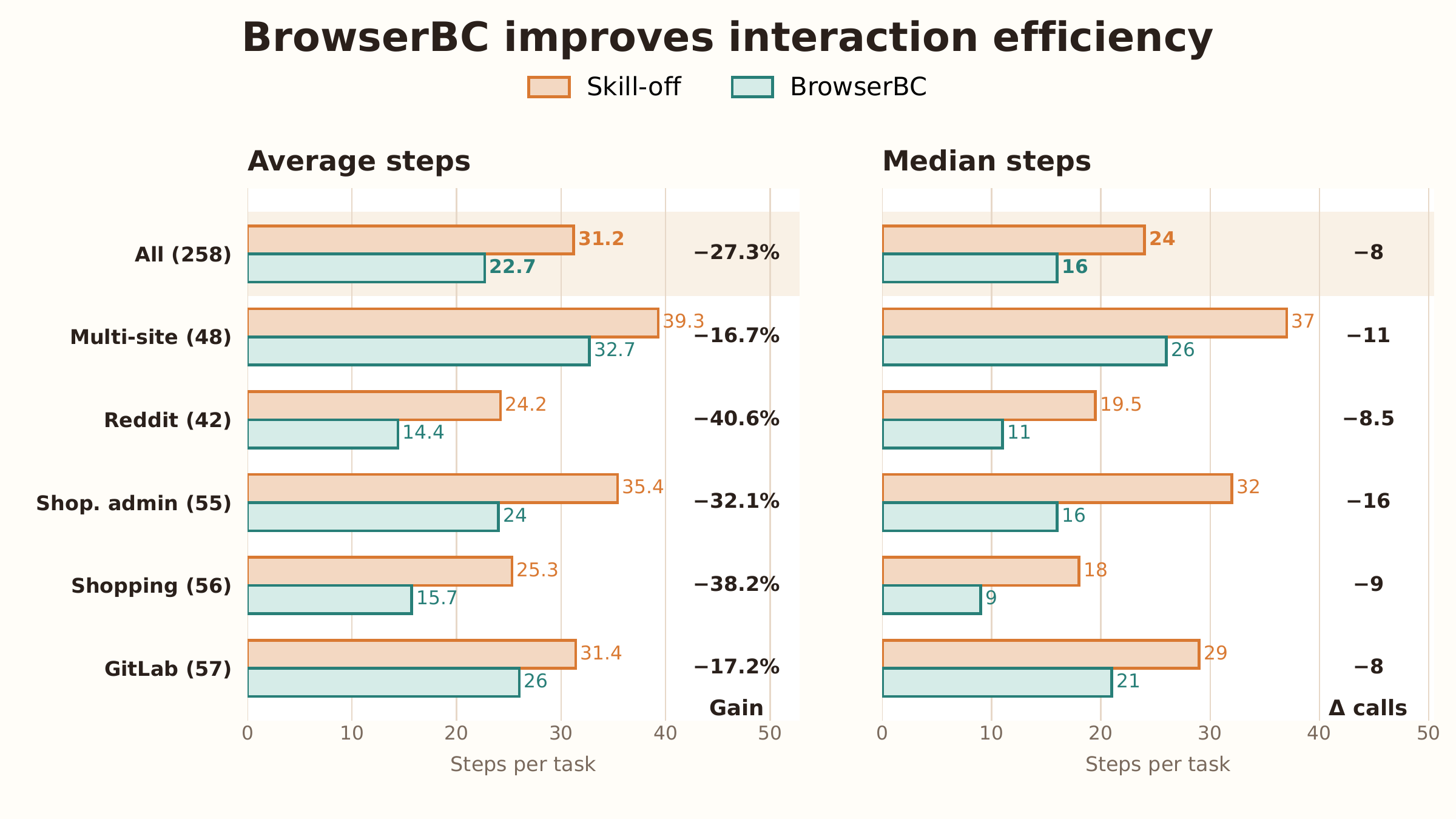}
    \caption{Interaction efficiency on WebArena-Hard by task group.}
    \label{fig:steps}
\end{subfigure}
\hfill
\begin{subfigure}[t]{0.40\linewidth}
    \centering
    \includegraphics[
        width=1.1111\figH,
        height=\figH
    ]{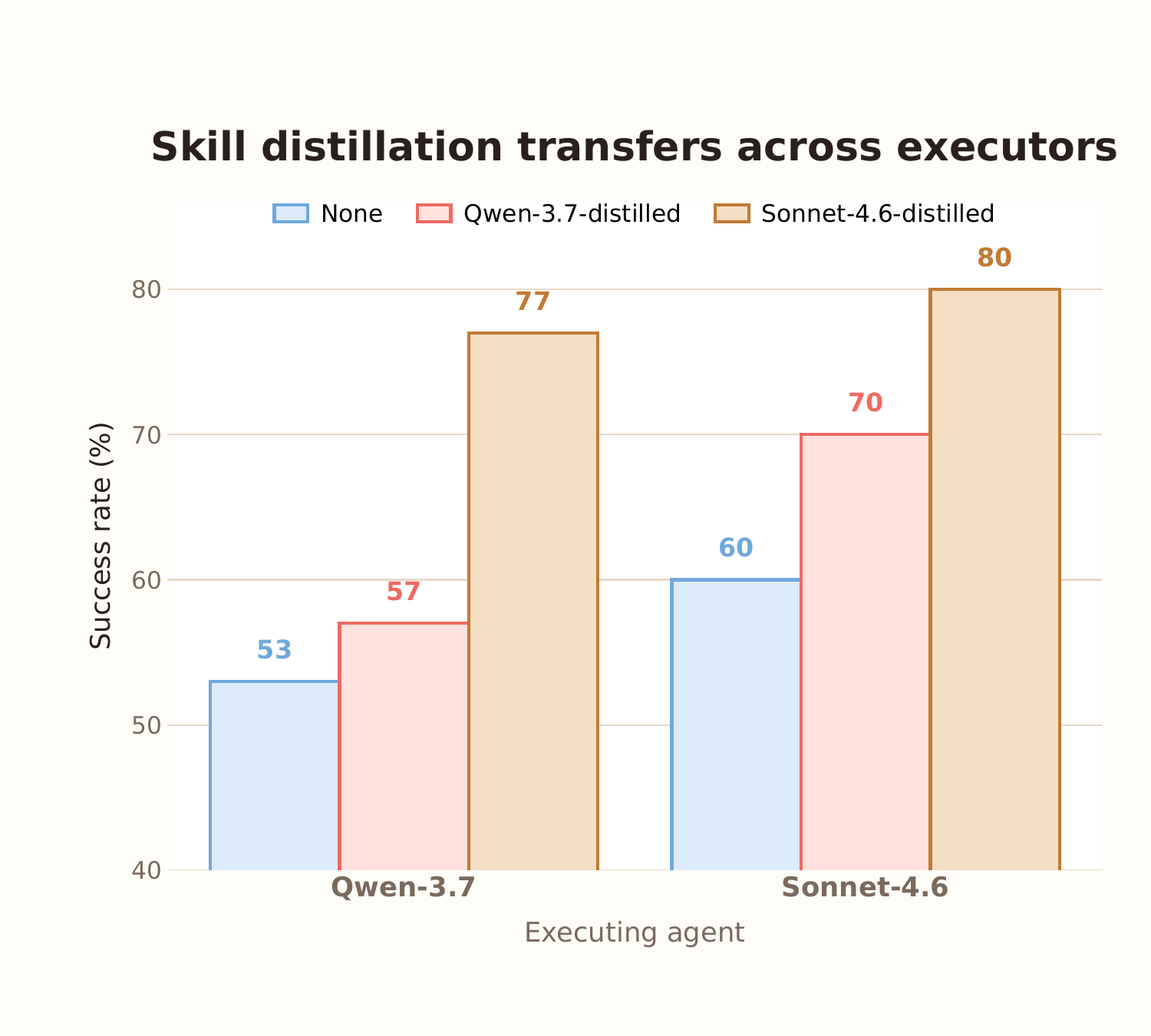}
    \caption{Decoupling the distiller from the executor.}
    \label{fig:distiller}
\end{subfigure}

\caption{\methodname{} improves interaction efficiency and transfers distilled skills across executors. Left: \methodname{} reduces the average number of steps in every group and lowers the overall median by $8$ tool calls. Right: on WebArena-Hard tasks, bars are grouped by the \emph{executing} agent and colored by the \emph{distilling} model.}
\label{fig:efficiency_and_transfer}
\end{figure}

\paragraph{Skills reduce interaction length.}
Beyond success rate, \methodname{} shortens the interaction required to solve a task. Across all WebArena-Hard tasks, the mean number of agent tool calls drops from $31.2$ to $22.7$ ($-27.3\%$) and the median from $24$ to $16$ ($-8$ steps), broken down by task group in Figure~\ref{fig:steps}. This reduction is consistent with the role of skills as procedural priors: they curtail exploratory navigation and repeated page inspection while leaving low-level grounding to the live browser state.

\paragraph{Decoupling the distiller from the executor.}
A central design claim of \methodname{} is that a skill is a model-agnostic object: it can be distilled once and reused by a different, cheaper agent at execution time (\S\ref{sec:formulation}). We test this by crossing the model that \emph{distills} skills with the model that \emph{executes} them on $30$ read-only WebArena-Hard tasks, using identical trajectories and the same evaluation protocol. We visualize the six resulting configurations as a grouped bar chart (Figure~\ref{fig:distiller}), grouping by executor and coloring by skill source, which renders the two effects directly comparable: \emph{within} each executor group the bars rise sharply only for Sonnet-distilled skills, while \emph{across} groups the Sonnet-distilled bars are of similar height. Two findings emerge. First, skill quality is determined primarily at distillation time: Sonnet-4.6-distilled skills improve both executors substantially ($+24$ and $+20$ points), whereas Qwen-3.7-plus-distilled skills yield only marginal gains. Second, high-quality skills transfer across executors: the small agent equipped with Sonnet-4.6-distilled skills reaches $77\%$, approaching the large agent's $80\%$, directly substantiating the ``distill once, reuse cheaply'' premise.

\paragraph{Skills are calibrated priors, not mandatory scripts.}
Finally, we examine whether retrieved skills should be followed unconditionally. An always-follow variant, which executes the retrieved skill without allowing the agent to override it against the live state, reaches only $77.5\%$ on WebArena-Hard, below \methodname{}'s $81.4\%$. A skill is thus best treated as a calibrated prior that the agent may revise when live observations contradict it, rather than a rigid program---a property we probe more sharply through the mismatched-skill controls in the desktop setting (\S\ref{sec:exp-osworld}).

\subsection{Beyond the Browser: OSWorld-Style Case Study}
\label{sec:exp-osworld}

\paragraph{Framing and protocol.}
The preceding experiments evaluate \methodname{} on browser tasks. We additionally include a small OSWorld-style desktop study to test a narrower question: whether the same skill object retains explanatory value when the interaction substrate changes from web pages to general GUI state. This study is not intended as an OSWorld leaderboard result or as an estimate of full OSWorld performance. Rather, it is a controlled case study for the paper's central representational claim: the reusable unit is not a click sequence, a DOM selector, or a browser-specific route, but a semantic procedure describing preconditions, state transitions, progress evidence, terminal evidence, and recovery.

We construct $30$ Ubuntu desktop cases covering settings changes, file and text editing, terminal commands, application configuration, modal dialogs, and office applications. Each case has a fixed initial state and an independent state-based verifier. Skills are distilled from successful demonstrations at the task-family level: a skill may describe that a setting must be opened, changed, persisted, and verified, or that a file must be edited and saved, but it does not encode coordinates or a replay script. We include cases where the skill matches the task, easy cases that the base agent already solves, and still-hard cases where the procedure is known but execution fails. This design lets the case study distinguish three effects that are difficult to separate in aggregate success rates: procedural uncertainty, execution fragility, and skill calibration.

\begin{figure}[t]
\centering
\begin{subfigure}[t]{\linewidth}
    \centering
    \includegraphics[width=\linewidth]{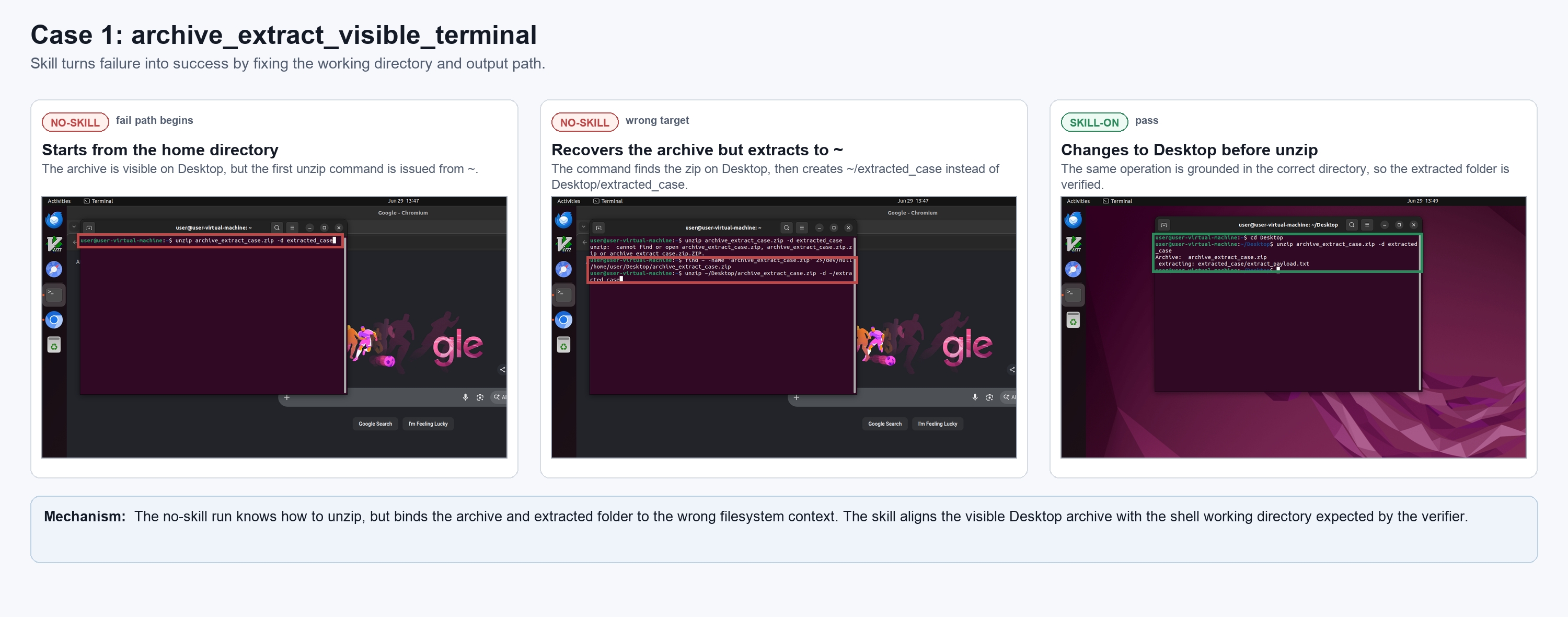}
\end{subfigure}

\vspace{0.5em}

\begin{subfigure}[t]{\linewidth}
    \centering
    \includegraphics[width=\linewidth]{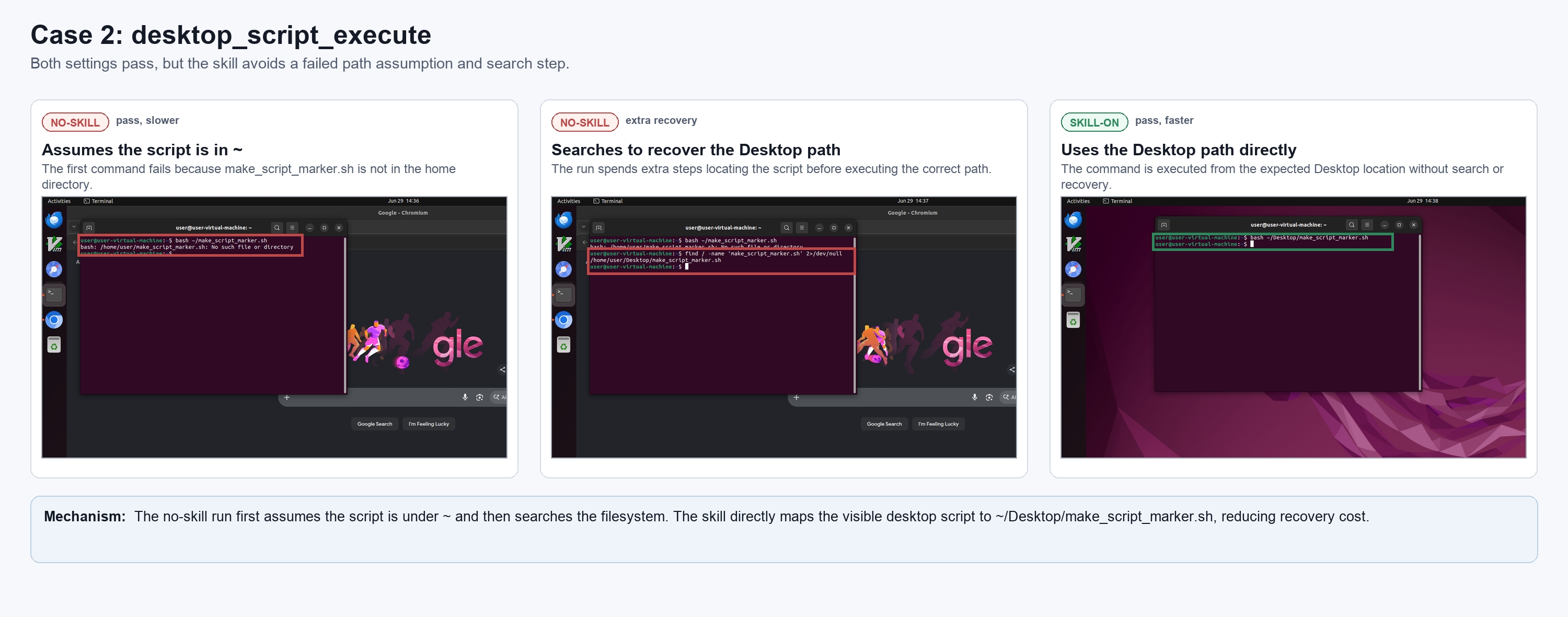}
\end{subfigure}

\caption{OSWorld-style desktop case studies. In both cases the no-skill agent
knows the correct command but binds it to the wrong filesystem context, while
the matched skill makes the implicit path-grounding procedure explicit.
}
\label{fig:osworld-cases}
\end{figure}

\begin{table}[ht]
\centering
\caption{Outcome taxonomy for the OSWorld-style diagnostic study. Counts denote cases in this controlled case set: 11 are solved without the skill, 6 additional cases are solved with the skill, and 13 remain unsolved in both settings.}
\label{tab:osworld-mechanisms}
\begin{adjustbox}{max width=0.85\linewidth}
\begin{tabular}{lcl}
\toprule
Case type & Count & Interpretation \\
\midrule
Solved without skill & $11$ & The base agent already finds the short action path. \\
Skill-enabled gain & $6$ & The skill resolves cases by supplying the missing task schema and terminal evidence. \\
Unsolved & $13$ & Failures persist due to focus, modal, file-picker, or recovery-state issues. \\
\bottomrule
\end{tabular}
\end{adjustbox}
\end{table}

\paragraph{Mechanistic interpretation.}

The positive cases show that skills transfer as \emph{task schemas}. In the archive extraction case (case 1 of Figure~\ref{fig:osworld-cases}), the no-skill agent already knows how to run the extraction command but fails to bind the visible Desktop archive to the verifier's expected output location. It first issues the command from the home directory, later recovers the archive path, but still creates the extracted folder under the wrong filesystem context. The matched skill repairs this failure by making the implicit procedure explicit: it establishes the Desktop working directory, performs the extraction there, and treats the resulting folder as terminal evidence---thereby aligning the visible archive with the shell working directory expected by the verifier.

The desktop script case (case 2 of Figure~\ref{fig:osworld-cases}) illustrates a second mechanism. Both settings ultimately pass, but the no-skill run first assumes that the script resides under the home directory and then spends extra steps searching the filesystem to recover the Desktop path. The matched skill maps the visible script directly to the Desktop shell path, eliminating the failed assumption and the subsequent recovery.


\paragraph{Scope.}
These cases also delineate the boundary of the method. Skill distillation can supply missing procedural structure, as in the archive extraction case, and can reduce avoidable recovery, as in the desktop script case. However, it is not a substitute for a shortest-path policy or for perfect low-level execution: in some cases the retrieved skill specifies a correct semantic routine, yet its extra verification step makes the successful trajectory slower than necessary. More generally, a skill should be treated as an evidence-weighted procedural prior that must be checked against the live GUI and filesystem state, not as an unconditional command. The case study therefore sharpens the central claim: distilled skills are portable when a task exhibits repeated procedural structure and verifiable intermediate states, but their benefit is bounded by execution reliability, retrieval correctness, and the cost of unnecessary verification.



\section{Conclusion}

In this work, we presented a framework that converts raw agent trajectories into a reusable skill library through three stages: Behavioral Evidence Abstraction, Procedural Skill Distillation, and Skill-Library Construction. By decoupling skill distillation from skill execution, our approach separates the question of \emph{what} constitutes a good skill from \emph{which} agent ultimately uses it, allowing skill quality to be fixed once at distillation time and transferred across heterogeneous executors. Experiments show that distilled skills consistently improve downstream task success and that stronger distillers yield skills that benefit even weaker executing agents.

These results indicate that explicitly distilling procedural skills from past trajectories is an effective way to accumulate and reuse experience across tasks and agents. We see several promising directions for future work, including extending the library to longer-horizon and multi-agent settings, and studying how skills can be continually refined and pruned as the agent encounters new environments.

\clearpage
\newpage
\definecolor{brandcolor}{RGB}{227, 170, 121}
\section{Contributions and Acknowledgements}

Zheng Jiang\textsuperscript{*}, Yuzhao Peng\textsuperscript{*}, Houde Qian\textsuperscript{*}, Kaisen Yang\textsuperscript{*,\dag}, Boshi Zhang\textsuperscript{*}, Youjie Zheng\textsuperscript{*}, Shijin Hong, Qingle Liu, Ruoyu Han, Bohan Lyu, Bingxiang He, Eren Cai, Calvin Xiao, Qinhuai Na\textsuperscript{\ddag}

\vspace{0.5em}
\noindent\textsuperscript{*}Equal contribution. Names are sorted in alphabetical order of the last name.\\
\textsuperscript{\dag}Project lead, \url{yks23@mails.tsinghua.edu.cn}.\\
\textsuperscript{\ddag}Corresponding author, \url{nana@einsia.ai}.

This project is founded and supported by Navers Lab, Einsia.AI.

\clearpage
\newpage
\bibliographystyle{assets/plainnat}
\bibliography{ref}

@inproceedings{prabhu2026walt,
  title        = {{WALT}: Web Agents that Learn Tools},
  author       = {Prabhu, Viraj and Dai, Yutong and Fernandez, Matthew and others},
  booktitle    = {International Conference on Learning Representations},
  year         = {2026},
  url          = {https://openreview.net/forum?id=cgIDqcJcoI}
}

@inproceedings{li2026websailorv2,
  title        = {WebSailor-V2: Bridging the Chasm to Proprietary Agents via Synthetic Data and Scalable Reinforcement Learning},
  author       = {Li, Kuan and Zhang, Zhongwang and Yin, Huifeng and others},
  booktitle    = {International Conference on Learning Representations},
  year         = {2026},
  url          = {https://openreview.net/forum?id=HuP16O5SJf}
}

@inproceedings{gandhi2026gobrowse,
  title        = {Go-Browse: Training Web Agents with Structured Exploration},
  author       = {Gandhi, Apurva and Neubig, Graham},
  booktitle    = {International Conference on Learning Representations},
  year         = {2026},
  url          = {https://openreview.net/forum?id=IpzRWE52yw}
}

@inproceedings{ye2026agentfold,
  title        = {AgentFold: Long-Horizon Web Agents with Proactive Context Folding},
  author       = {Ye, Rui and Zhang, Zhongwang and Li, Kuan and others},
  booktitle    = {International Conference on Learning Representations},
  year         = {2026},
  url          = {https://openreview.net/forum?id=IuZoTgsUws}
}

@inproceedings{liu2026scalecua,
  title        = {ScaleCUA: Scaling Open-Source Computer Use Agents with Cross-Platform Data},
  author       = {Liu, Zhaoyang and Xie, JingJing and Ding, Zichen and others},
  booktitle    = {International Conference on Learning Representations},
  year         = {2026},
  url          = {https://openreview.net/forum?id=yBFUqdJFZn}
}

@inproceedings{chae2025webshepherd,
  title        = {Web-Shepherd: Advancing {PRMs} for Reinforcing Web Agents},
  author       = {Chae, Hyungjoo and Kim, Sunghwan and Cho, Junhee and others},
  booktitle    = {Advances in Neural Information Processing Systems},
  year         = {2025},
  url          = {https://openreview.net/forum?id=G2kMroO9UV}
}

@inproceedings{wu2025webdancer,
  title        = {WebDancer: Towards Autonomous Information Seeking Agency},
  author       = {Wu, Jialong and Li, Baixuan and Fang, Runnan and others},
  booktitle    = {Advances in Neural Information Processing Systems},
  year         = {2025},
  url          = {https://openreview.net/forum?id=quJdphBcdP}
}

@inproceedings{yang2025sage,
  title        = {Self-Guided Hierarchical Exploration for Generalist Foundation Model Web Agents},
  author       = {Yang, Qianlan and Wang, Xiangjun and Perszyk, Danielle and Wang, Yu-Xiong},
  booktitle    = {Advances in Neural Information Processing Systems},
  year         = {2025},
  url          = {https://openreview.net/forum?id=9twwDW60Bw}
}

@inproceedings{vattikonda2025trainwebagent,
  title        = {How to Train Your {LLM} Web Agent: A Statistical Diagnosis},
  author       = {Vattikonda, Dheeraj and Ravichandran, Santhoshi and Penaloza, Emiliano and others},
  booktitle    = {Advances in Neural Information Processing Systems},
  year         = {2025},
  url          = {https://openreview.net/forum?id=67xkPEM3bZ}
}

@inproceedings{shen2025thinkingdoing,
  title        = {Thinking vs. Doing: Improving Agent Reasoning by Scaling Test-Time Interaction},
  author       = {Shen, Junhong and Bai, Hao and Zhang, Lunjun and others},
  booktitle    = {Advances in Neural Information Processing Systems},
  year         = {2025},
  url          = {https://openreview.net/forum?id=un1TRwNgiv}
}

@inproceedings{wang2025opencua,
  title        = {OpenCUA: Open Foundations for Computer-Use Agents},
  author       = {Wang, Xinyuan and Wang, Bowen and Lu, Dunjie and others},
  booktitle    = {Advances in Neural Information Processing Systems},
  year         = {2025},
  url          = {https://openreview.net/forum?id=6iRZvJiC9Q}
}

@misc{yao2023webshopscalablerealworldweb,
      title={WebShop: Towards Scalable Real-World Web Interaction with Grounded Language Agents}, 
      author={Shunyu Yao and Howard Chen and John Yang and Karthik Narasimhan},
      year={2023},
      eprint={2207.01206},
      archivePrefix={arXiv},
      primaryClass={cs.CL},
      url={https://arxiv.org/abs/2207.01206}, 
}

@misc{deng2023mind2webgeneralistagentweb,
      title={Mind2Web: Towards a Generalist Agent for the Web}, 
      author={Xiang Deng and Yu Gu and Boyuan Zheng and Shijie Chen and Samuel Stevens and Boshi Wang and Huan Sun and Yu Su},
      year={2023},
      eprint={2306.06070},
      archivePrefix={arXiv},
      primaryClass={cs.CL},
      url={https://arxiv.org/abs/2306.06070}, 
}

@misc{zhou2024webarenarealisticwebenvironment,
      title={WebArena: A Realistic Web Environment for Building Autonomous Agents}, 
      author={Shuyan Zhou and Frank F. Xu and Hao Zhu and Xuhui Zhou and Robert Lo and Abishek Sridhar and Xianyi Cheng and Tianyue Ou and Yonatan Bisk and Daniel Fried and Uri Alon and Graham Neubig},
      year={2024},
      eprint={2307.13854},
      archivePrefix={arXiv},
      primaryClass={cs.AI},
      url={https://arxiv.org/abs/2307.13854}, 
}

@misc{koh2024visualwebarenaevaluatingmultimodalagents,
      title={VisualWebArena: Evaluating Multimodal Agents on Realistic Visual Web Tasks}, 
      author={Jing Yu Koh and Robert Lo and Lawrence Jang and Vikram Duvvur and Ming Chong Lim and Po-Yu Huang and Graham Neubig and Shuyan Zhou and Ruslan Salakhutdinov and Daniel Fried},
      year={2024},
      eprint={2401.13649},
      archivePrefix={arXiv},
      primaryClass={cs.LG},
      url={https://arxiv.org/abs/2401.13649}, 
}

@misc{lù2024weblinxrealworldwebsitenavigation,
      title={WebLINX: Real-World Website Navigation with Multi-Turn Dialogue}, 
      author={Xing Han Lù and Zdeněk Kasner and Siva Reddy},
      year={2024},
      eprint={2402.05930},
      archivePrefix={arXiv},
      primaryClass={cs.CL},
      url={https://arxiv.org/abs/2402.05930}, 
}

@misc{drouin2024workarenacapablewebagents,
      title={WorkArena: How Capable Are Web Agents at Solving Common Knowledge Work Tasks?}, 
      author={Alexandre Drouin and Maxime Gasse and Massimo Caccia and Issam H. Laradji and Manuel Del Verme and Tom Marty and Léo Boisvert and Megh Thakkar and Quentin Cappart and David Vazquez and Nicolas Chapados and Alexandre Lacoste},
      year={2024},
      eprint={2403.07718},
      archivePrefix={arXiv},
      primaryClass={cs.LG},
      url={https://arxiv.org/abs/2403.07718}, 
}

@misc{boisvert2025workarenacompositionalplanningreasoningbased,
      title={WorkArena++: Towards Compositional Planning and Reasoning-based Common Knowledge Work Tasks}, 
      author={Léo Boisvert and Megh Thakkar and Maxime Gasse and Massimo Caccia and Thibault Le Sellier De Chezelles and Quentin Cappart and Nicolas Chapados and Alexandre Lacoste and Alexandre Drouin},
      year={2025},
      eprint={2407.05291},
      archivePrefix={arXiv},
      primaryClass={cs.AI},
      url={https://arxiv.org/abs/2407.05291}, 
}

@misc{dechezelles2025browsergymecosystemwebagent,
      title={The BrowserGym Ecosystem for Web Agent Research}, 
      author={Thibault Le Sellier De Chezelles and Maxime Gasse and Alexandre Drouin and Massimo Caccia and Léo Boisvert and Megh Thakkar and Tom Marty and Rim Assouel and Sahar Omidi Shayegan and Lawrence Keunho Jang and Xing Han Lù and Ori Yoran and Dehan Kong and Frank F. Xu and Siva Reddy and Quentin Cappart and Graham Neubig and Ruslan Salakhutdinov and Nicolas Chapados and Alexandre Lacoste},
      year={2025},
      eprint={2412.05467},
      archivePrefix={arXiv},
      primaryClass={cs.LG},
      url={https://arxiv.org/abs/2412.05467}, 
}

@misc{gur2024realworldwebagentplanninglong,
      title={A Real-World WebAgent with Planning, Long Context Understanding, and Program Synthesis}, 
      author={Izzeddin Gur and Hiroki Furuta and Austin Huang and Mustafa Safdari and Yutaka Matsuo and Douglas Eck and Aleksandra Faust},
      year={2024},
      eprint={2307.12856},
      archivePrefix={arXiv},
      primaryClass={cs.LG},
      url={https://arxiv.org/abs/2307.12856}, 
}

@misc{zheng2024gpt4visiongeneralistwebagent,
      title={GPT-4V(ision) is a Generalist Web Agent, if Grounded}, 
      author={Boyuan Zheng and Boyu Gou and Jihyung Kil and Huan Sun and Yu Su},
      year={2024},
      eprint={2401.01614},
      archivePrefix={arXiv},
      primaryClass={cs.IR},
      url={https://arxiv.org/abs/2401.01614}, 
}

@misc{he2024webvoyagerbuildingendtoendweb,
      title={WebVoyager: Building an End-to-End Web Agent with Large Multimodal Models}, 
      author={Hongliang He and Wenlin Yao and Kaixin Ma and Wenhao Yu and Yong Dai and Hongming Zhang and Zhenzhong Lan and Dong Yu},
      year={2024},
      eprint={2401.13919},
      archivePrefix={arXiv},
      primaryClass={cs.CL},
      url={https://arxiv.org/abs/2401.13919}, 
}

@misc{he2024openwebvoyagerbuildingmultimodalweb,
      title={OpenWebVoyager: Building Multimodal Web Agents via Iterative Real-World Exploration, Feedback and Optimization}, 
      author={Hongliang He and Wenlin Yao and Kaixin Ma and Wenhao Yu and Hongming Zhang and Tianqing Fang and Zhenzhong Lan and Dong Yu},
      year={2024},
      eprint={2410.19609},
      archivePrefix={arXiv},
      primaryClass={cs.CL},
      url={https://arxiv.org/abs/2410.19609}, 
}

@misc{xie2024osworldbenchmarkingmultimodalagents,
      title={OSWorld: Benchmarking Multimodal Agents for Open-Ended Tasks in Real Computer Environments}, 
      author={Tianbao Xie and Danyang Zhang and Jixuan Chen and Xiaochuan Li and Siheng Zhao and Ruisheng Cao and Toh Jing Hua and Zhoujun Cheng and Dongchan Shin and Fangyu Lei and Yitao Liu and Yiheng Xu and Shuyan Zhou and Silvio Savarese and Caiming Xiong and Victor Zhong and Tao Yu},
      year={2024},
      eprint={2404.07972},
      archivePrefix={arXiv},
      primaryClass={cs.AI},
      url={https://arxiv.org/abs/2404.07972}, 
}

@misc{shinn2023reflexionlanguageagentsverbal,
      title={Reflexion: Language Agents with Verbal Reinforcement Learning}, 
      author={Noah Shinn and Federico Cassano and Edward Berman and Ashwin Gopinath and Karthik Narasimhan and Shunyu Yao},
      year={2023},
      eprint={2303.11366},
      archivePrefix={arXiv},
      primaryClass={cs.AI},
      url={https://arxiv.org/abs/2303.11366}, 
}

@misc{zhao2024expelllmagentsexperiential,
      title={ExpeL: LLM Agents Are Experiential Learners}, 
      author={Andrew Zhao and Daniel Huang and Quentin Xu and Matthieu Lin and Yong-Jin Liu and Gao Huang},
      year={2024},
      eprint={2308.10144},
      archivePrefix={arXiv},
      primaryClass={cs.LG},
      url={https://arxiv.org/abs/2308.10144}, 
}

@misc{wang2023voyageropenendedembodiedagent,
      title={Voyager: An Open-Ended Embodied Agent with Large Language Models}, 
      author={Guanzhi Wang and Yuqi Xie and Yunfan Jiang and Ajay Mandlekar and Chaowei Xiao and Yuke Zhu and Linxi Fan and Anima Anandkumar},
      year={2023},
      eprint={2305.16291},
      archivePrefix={arXiv},
      primaryClass={cs.AI},
      url={https://arxiv.org/abs/2305.16291}, 
}

@misc{wang2025inducingprogrammaticskillsagentic,
      title={Inducing Programmatic Skills for Agentic Tasks}, 
      author={Zora Zhiruo Wang and Apurva Gandhi and Graham Neubig and Daniel Fried},
      year={2025},
      eprint={2504.06821},
      archivePrefix={arXiv},
      primaryClass={cs.CL},
      url={https://arxiv.org/abs/2504.06821}, 
}

@misc{yin2024agentlumosunifiedmodular,
      title={Agent Lumos: Unified and Modular Training for Open-Source Language Agents}, 
      author={Da Yin and Faeze Brahman and Abhilasha Ravichander and Khyathi Chandu and Kai-Wei Chang and Yejin Choi and Bill Yuchen Lin},
      year={2024},
      eprint={2311.05657},
      archivePrefix={arXiv},
      primaryClass={cs.AI},
      url={https://arxiv.org/abs/2311.05657}, 
}

@misc{xu2025agenttrekagenttrajectorysynthesis,
      title={AgentTrek: Agent Trajectory Synthesis via Guiding Replay with Web Tutorials}, 
      author={Yiheng Xu and Dunjie Lu and Zhennan Shen and Junli Wang and Zekun Wang and Yuchen Mao and Caiming Xiong and Tao Yu},
      year={2025},
      eprint={2412.09605},
      archivePrefix={arXiv},
      primaryClass={cs.CL},
      url={https://arxiv.org/abs/2412.09605}, 
}

@misc{gao2024memorysharing,
  title        = {Memory Sharing for Large Language Model based Agents},
  author       = {Gao, Hang and Zhang, Yongfeng},
  year         = {2024},
  eprint       = {2404.09982},
  archivePrefix = {arXiv},
  primaryClass = {cs.AI},
  url          = {https://arxiv.org/abs/2404.09982}
}

@inproceedings{openxembodiment2024,
  title        = {Open {X-Embodiment}: Robotic Learning Datasets and {RT-X} Models},
  author       = {{Open X-Embodiment Collaboration} and O'Neill, Abby and Rehman, Abdul and Maddukuri, Abhiram and Gupta, Abhishek and Padalkar, Abhishek and Lee, Abraham and Pooley, Acorn and others},
  booktitle    = {2024 IEEE International Conference on Robotics and Automation},
  pages        = {6892--6903},
  year         = {2024},
  doi          = {10.1109/ICRA57147.2024.10611477},
  url          = {https://arxiv.org/abs/2310.08864}
}

\clearpage
\newpage
\appendix
\clearpage
\onecolumn
\section{Appendix}
\addcontentsline{toc}{section}{Appendix}
\label{sec:appendix}

\begin{tcolorbox}[
  enhanced,
  colback=white,
  colframe=brandprimary,
  boxrule=0.6pt,
  arc=2pt,
  title={\sffamily\bfseries Appendix guide},
  fonttitle=\sffamily\bfseries,
  left=8pt,right=8pt,top=8pt,bottom=8pt
]
This appendix provides the implementation protocol, representation sketches, case
demonstrations, and per-task result analysis behind the aggregate results in
the main text.  Tables follow the same visual grammar as Section~\ref{sec:experiments}:
booktabs rules, brand-accent group headers, compact spacing, and highlighted
total rows.
\end{tcolorbox}

\subsection{Implementation Details}
\label{app:implementation}

\subsubsection{Skill Object Format}
\label{app:skill-format}

Across the web experiments, skills are represented as structured natural-language
cards rather than as executable programs or replay traces. Each card specifies the
task family it applies to, its preconditions, a concise procedure, progress and
completion evidence, common failure modes, recovery or abort rules, and
provenance. This structure is intentionally redundant with the agent's live
observations: the card supplies procedural priors, whereas the browser state
remains the source of truth for concrete labels, element identities, page
availability, validation messages, and final submission state.

In WebArena-Hard, this format is specialized to self-hosted browser tasks with
programmatic checkers.  A skill records the task pattern, the relevant site
state, the sequence of semantic operations that tends to make progress, and the
observable evidence required before stopping.  For example, a skill may say to
construct a report with a particular aggregation period and status filter, or to
verify a committed file state, but it does not prescribe fixed coordinates or a
fixed action sequence.

In ClawBench, the same skill format is adapted to live websites where terminal
states may be visible in the page, reflected in a submitted request, or judged by
an agentic evaluator.  The skill therefore emphasizes state preservation,
pre-submission checks, false terminal states, and recovery order.  Across both
benchmarks, skills are rendered as agent-facing procedural guidance: they state
which target state must be established and how completion should be verified,
without copying stale coordinates, DOM selectors, account tokens, verification
codes, or other private runtime values.

\subsubsection{WebArena-Hard Harness}
\label{app:webarena-harness}

WebArena-Hard is evaluated in the benchmark's self-hosted environment using the
original task definitions and programmatic checkers. Unless stated otherwise in
the main text, the skill-off and skill-on conditions use the same executing
agent, browser action interface, authentication state, and step budget; the only
intervention is whether learned skill guidance is provided.

The WebArena-Hard skill construction protocol is task-local.  Skills are
distilled only from demonstrations associated with the same task family, so
evidence from one benchmark task is not exposed to unrelated tasks.  During
execution, a retrieved skill is appended as procedural guidance rather than as a
mandatory plan: the prompt explicitly instructs the agent to trust the live
browser observation whenever it conflicts with the skill.  This is important in
longer administrative tasks, where a useful skill may provide the correct report
construction recipe while the page layout or intermediate filter state still
has to be grounded at run time.

\subsubsection{ClawBench Harness}
\label{app:clawbench-harness}

ClawBench is evaluated on live production websites with a fresh browser state
per case.  The results in Table~\ref{tab:main-results} use the final 152-case
evaluation set and aggregate outcomes by task category.  Because the websites
are live, the evaluation emphasizes robust procedural guidance rather than
memorization of a fixed page state.

The skill-on pipeline is iterative. Cases first run without a skill library;
successful demonstrations are then distilled into reusable skills.  On later
runs, each case receives only the matched case-level or domain-level skill, not
the full library.  This one-skill-per-case intervention makes the comparison a
test of targeted procedural priors rather than a broad memory dump.

The distillation step draws on the benchmark task description, evaluation
rubric, verified human demonstrations, and observable completion evidence when
available.  The resulting skills emphasize state preservation and terminal
evidence. For instance, if a successful human path preserved upstream search
filters or booking constraints before opening a final contact, checkout,
creation, or submission page, the skill renders those constraints as a
pre-terminal gate that must be rechecked in the live browser before final
submission.

\subsubsection{Evaluation and Leakage Controls}
\label{app:evaluation-controls}

WebArena-Hard uses the benchmark's original programmatic checkers, whereas
ClawBench uses an agentic pass/fail evaluator that applies the benchmark rubric
to the completed trajectory.  The evaluator records its justification, the
rubric rules applied, and the key supporting evidence for audit.

All skill-off and skill-on comparisons hold the executing agent, action interface, and step budget fixed, varying only whether learned skill guidance is available. Skills are restricted to procedural content: they may describe what state to establish, which task-provided values must be preserved, what progress looks like, and what terminal evidence confirms completion, but they may not encode checker-only answers or stale runtime literals. Private or volatile values, such as account tokens, verification codes, payment details, generated identifiers, and session-specific contact information, are redacted or treated as non-copyable runtime values.

\section{Trajectory and Skill Representations}
\label{app:webarena-artifacts}

This appendix visualizes the representations used by \methodname{} and gives concrete
WebArena case demonstrations.  A demonstration begins as a human browser
trajectory, is distilled into a reusable Markdown skill, and is then evaluated
through a fresh agent rollout.  The figures and boxes below use the same palette
as the main paper.

\paragraph{Representation sketches.}
The appendix exposes both the normalized trajectory view and the
distilled skill view.  The former preserves task evidence; the latter abstracts
that evidence into reusable procedural guidance.

\noindent
\begin{minipage}[t]{0.47\linewidth}
\begin{tcolorbox}[
  enhanced,
  colback=white,
  colframe=brandprimary,
  boxrule=0.6pt,
  arc=2pt,
  title={\sffamily\bfseries Demonstration card},
  fonttitle=\sffamily\bfseries,
  left=8pt,right=8pt,top=8pt,bottom=8pt
]
{\ttfamily\footnotesize
\textcolor{brandaccent}{\#\# Metadata}\\
site: \textless{}site\textgreater{}\\
intent: \textless{}user request\textgreater{}\\[4pt]
\textcolor{brandaccent}{\#\# Step sequence}\\
- url: \textless{}page\textgreater{}\\
\hspace*{1em}observation: \textless{}visible state\textgreater{}\\
\hspace*{1em}action: \textless{}browser action\textgreater{}\\
\hspace*{1em}evidence: \textless{}why it matters\textgreater{}\\[4pt]
\textcolor{brandaccent}{\#\# Stop record}\\
answer: \textless{}result\textgreater{}\\
stop\_signal: \textless{}checked evidence\textgreater{}
}
\end{tcolorbox}
\end{minipage}
\hfill
\begin{minipage}[t]{0.47\linewidth}
\begin{tcolorbox}[
  enhanced,
  colback=white,
  colframe=brandprimary,
  boxrule=0.6pt,
  arc=2pt,
  title={\sffamily\bfseries Skill card},
  fonttitle=\sffamily\bfseries,
  left=8pt,right=8pt,top=8pt,bottom=8pt
]
{\ttfamily\footnotesize
\textcolor{brandaccent}{\#\# Metadata}\\
pattern: \textless{}task pattern\textgreater{}\\
goal: \textless{}what to accomplish\textgreater{}\\[4pt]
\textcolor{brandaccent}{\#\# Procedure}\\
- precondition: \textless{}site state\textgreater{}\\
\hspace*{1em}navigate: \textless{}target page\textgreater{}\\
\hspace*{1em}act: \textless{}browser action\textgreater{}\\
\hspace*{1em}verify: \textless{}why it works\textgreater{}\\[4pt]
\textcolor{brandaccent}{\#\# Stop rule}\\
answer: \textless{}result type\textgreater{}\\
stop\_signal: \textless{}completion evidence\textgreater{}
}
\end{tcolorbox}
\end{minipage}

\vspace{0.6em}

\section{WebArena Case Demonstrations}
\label{app:webarena-case-demos}

The case demonstrations compare a base rollout with a skill-conditioned rollout
for the same task.  Each panel uses real browser screenshots from the recorded
rollout and a short caption describing the behavior encoded by the trajectory or
skill.

\begin{figure}[t]
\centering
\includegraphics[width=0.96\linewidth]{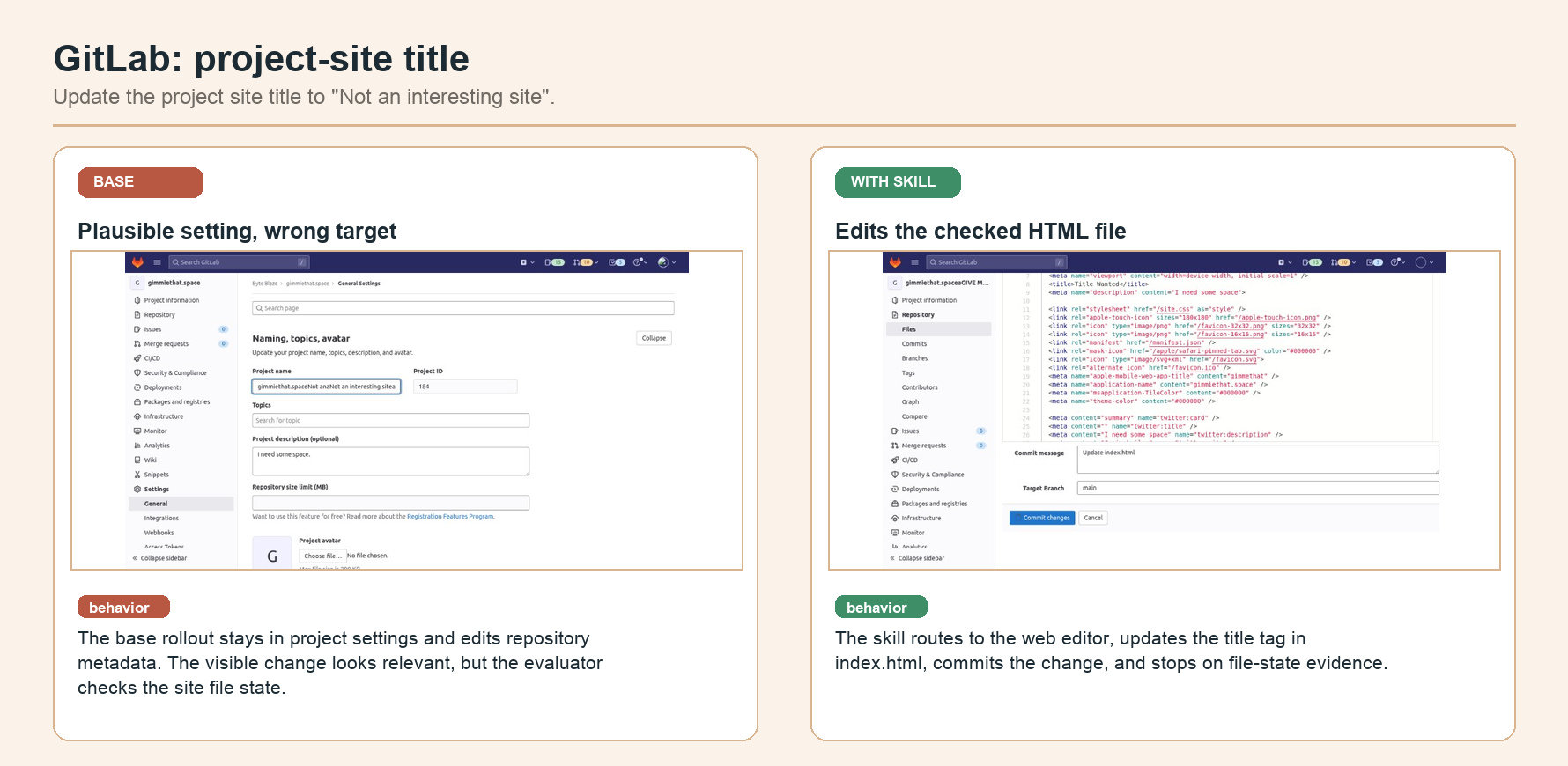}
\caption{GitLab case demo.  The base rollout changes a plausible project
setting, while the skill-conditioned rollout edits the checked HTML file and
commits the target state.}
\label{fig:appendix-case-gitlab}
\end{figure}

\begin{figure}[t]
\centering
\includegraphics[width=0.96\linewidth]{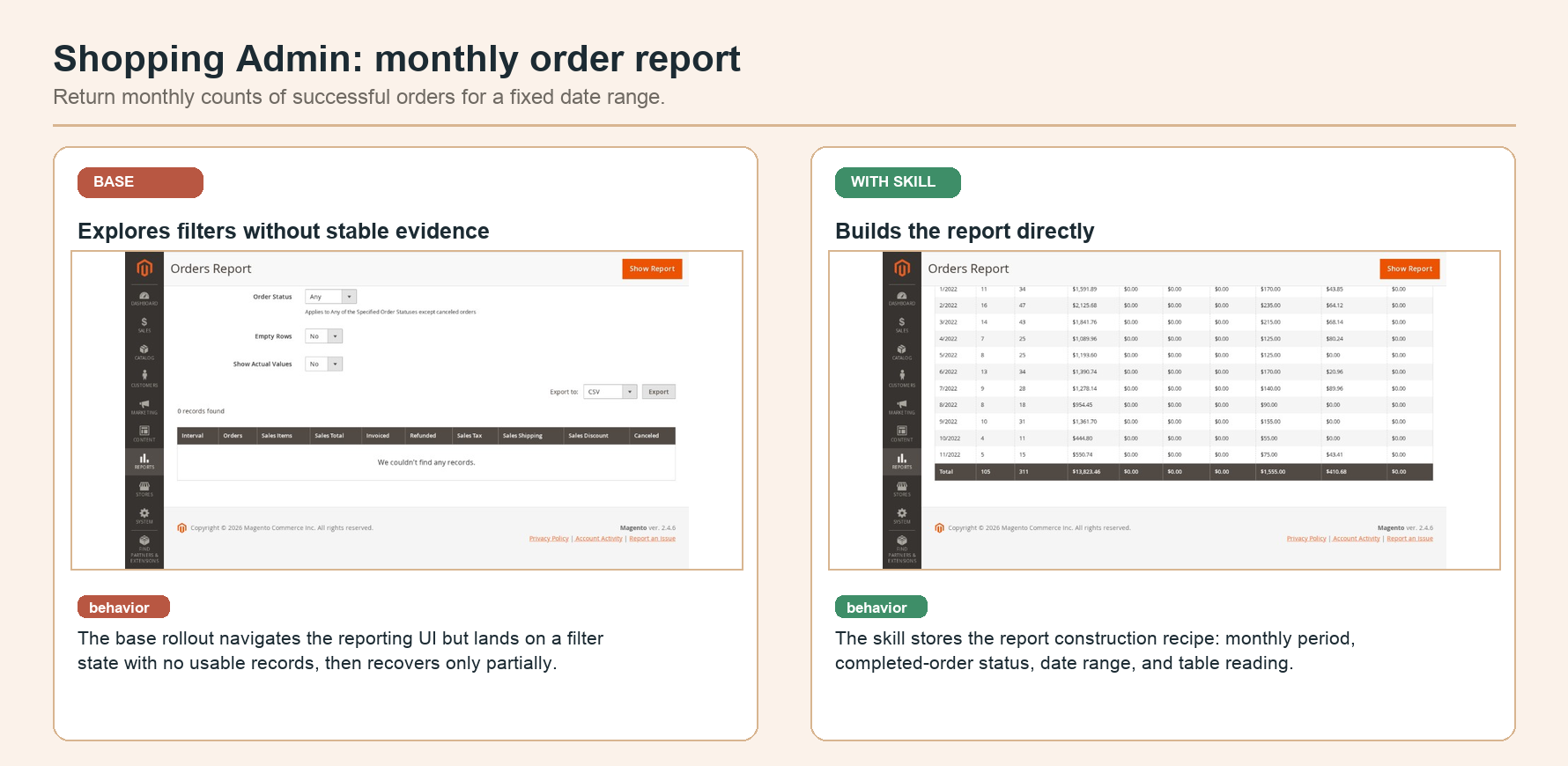}
\caption{Shopping Admin case demo.  The skill captures the report-construction
procedure: date range, monthly aggregation, completed-order filter, and table
reading.}
\label{fig:appendix-case-admin}
\end{figure}

\begin{figure}[t]
\centering
\includegraphics[width=0.96\linewidth]{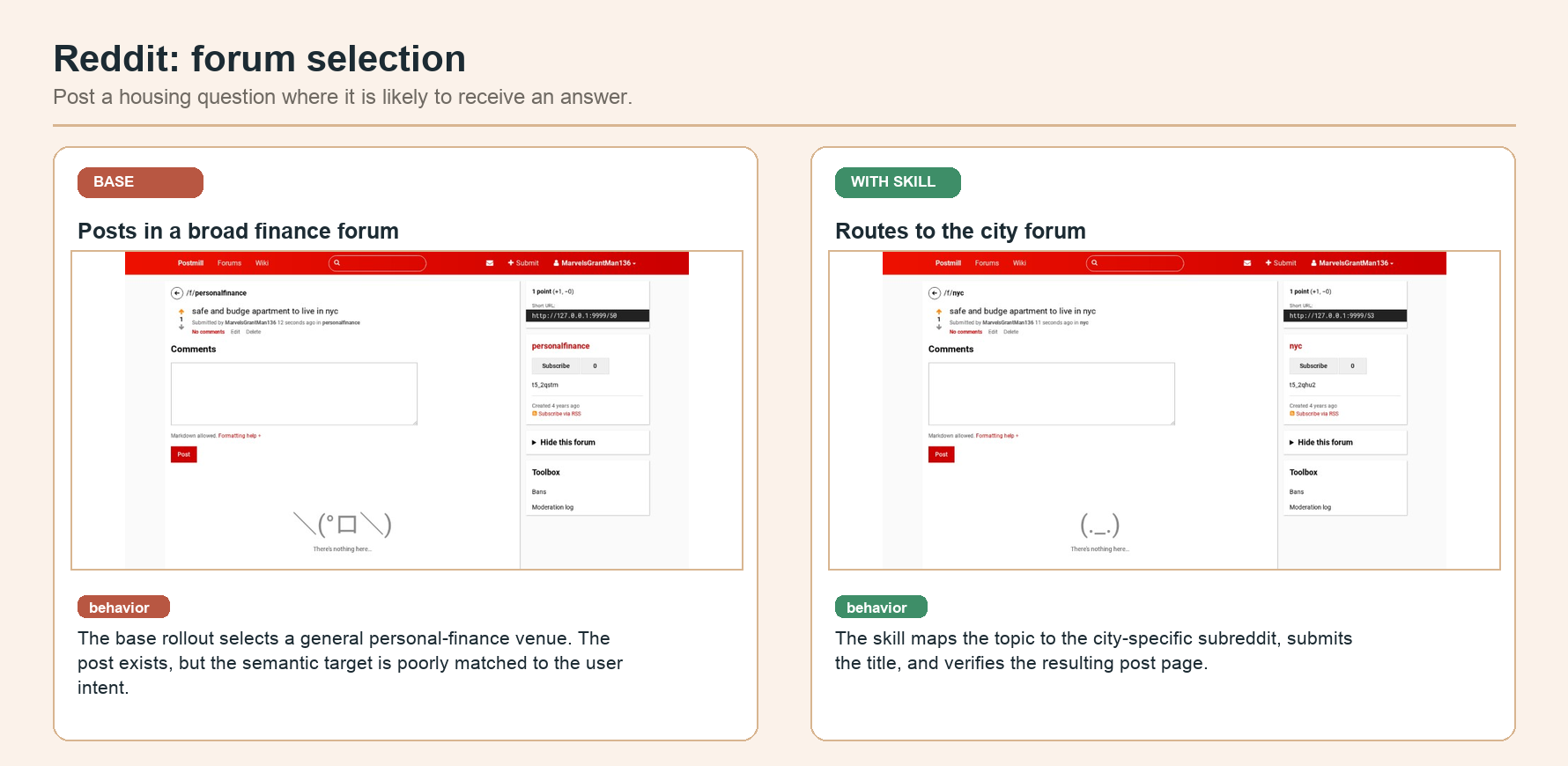}
\caption{Reddit case demo.  The skill encodes semantic forum selection and
routes the housing question to the city-specific forum before verifying the
created post.}
\label{fig:appendix-case-reddit}
\end{figure}

\begin{figure}[t]
\centering
\includegraphics[width=0.96\linewidth]{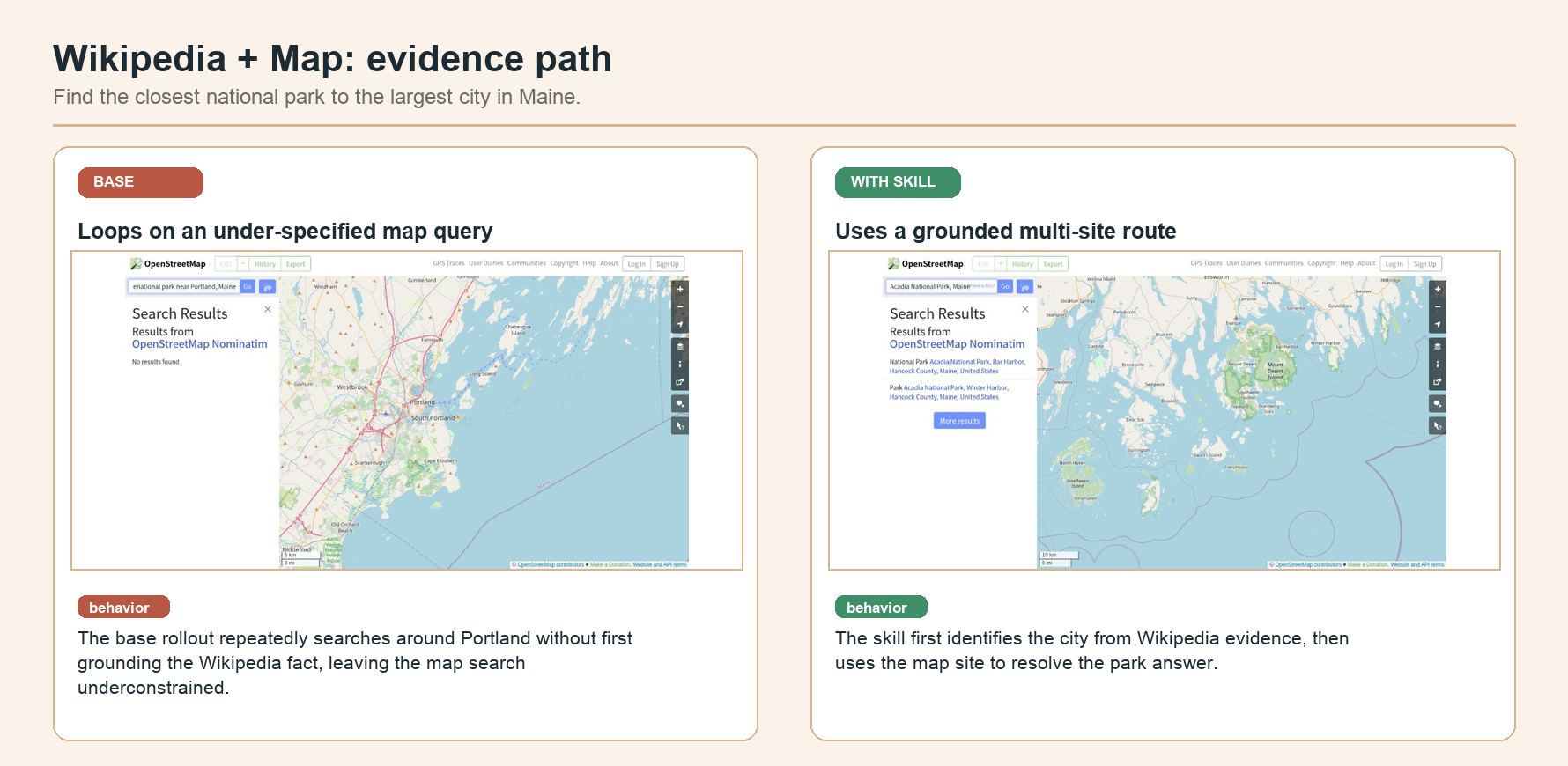}
\caption{Wikipedia and map case demo.  The skill preserves the cross-site
evidence order: ground the city fact first, then query the map site for the
park answer.}
\label{fig:appendix-case-map}
\end{figure}
\subsection{Additional Results}
\label{sec:appendix-results}

This section summarizes the scoring and interaction-efficiency checks behind the
WebArena-Hard aggregate reported in the main text.

\paragraph{Audited scoring.}
The final WebArena-Hard totals are 156 / 258 solved tasks for the skill-off
baseline and 210 / 258 solved tasks for \methodname{}, corresponding to 60.5\%
and 81.4\% success respectively.  Manual adjudication, when required, is applied
before aggregation under the same pass/fail standard used by the benchmark
checker, so the appendix reports one audited outcome per task.

\paragraph{Interaction efficiency.}
The per-task records also support the interaction-efficiency analysis in
Figure~\ref{fig:steps}: the all-task median decreases from 24 to 16 tool calls,
and every WebArena-Hard group shows shorter trajectories under skill
conditioning.  This pattern is consistent with \methodname{} acting as a
procedural prior: retrieved skills reduce exploratory navigation and repeated
inspection while leaving concrete grounding to the live browser state.

\paragraph{Qualitative appendix material.}
Case-level qualitative material uses the same outcome labels as the per-task
breakdown, making individual successes and failures comparable to the aggregate
tables.

\begin{table}[t]
\centering
\caption{Qualitative appendix material linked to the per-task breakdown.}
\label{tab:appendix-qualitative-material}
\renewcommand{\arraystretch}{1.08}
\setlength{\tabcolsep}{10pt}
\footnotesize
\begin{adjustbox}{max width=\linewidth}
\begin{tabular}{lll}
\toprule
Material & Scope & Fields \\
\midrule
\multicolumn{3}{l}{\sffamily\bfseries\color{brandaccent}Case-level evidence} \\
\cmidrule(lr){1-3}
OSWorld-style case set & Desktop cases & Task family, case type, verifier, skill-off/on outcome. \\
Qualitative browser examples & WebArena and ClawBench & Matched skill, agent actions, terminal evidence. \\
Failure analysis & Browser and desktop failures & Execution fragility, retrieval miss, over-following, recovery. \\
\bottomrule
\end{tabular}
\end{adjustbox}
\end{table}

\end{document}